\documentclass{article}

\usepackage{arxiv}

\usepackage[utf8]{inputenc} 
\usepackage[T1]{fontenc}    
\usepackage{hyperref}       
\usepackage{url}            
\usepackage{booktabs}       
\usepackage{amsfonts}       
\usepackage{nicefrac}       
\usepackage{microtype}      
\usepackage{lipsum}
\usepackage{graphicx}
\usepackage{amsmath}
\usepackage{amssymb}
\graphicspath{ {./images/} }

\title{Chain of Questions: Guiding Multimodal Curiosity in Language Models}

\author{
 Nima Iji \\
  Edinburgh Napier University\\
  Edinburgh, UK\\
  \texttt{40656795@napier.ac.uk} \\
   \And
 Kia Dashtipour \\
  Edinburgh Napier University\\
  Edinburgh, UK\\
  \texttt{k.dashtipour@napier.ac.uk} \\
}

\begin{document}
\maketitle
\begin{abstract}
Reasoning capabilities in large language models (LLMs) have substantially advanced through methods such as chain-of-thought and explicit step-by-step explanations. However, these improvements have not yet fully transitioned to multimodal contexts, where models must proactively decide which sensory modalities such as vision, audio, or spatial perception to engage when interacting with complex real-world environments. In this paper, we introduce the Chain of Questions (CoQ) framework, a curiosity-driven reasoning approach that encourages multimodal language models to dynamically generate targeted questions regarding their surroundings. These generated questions guide the model to selectively activate relevant modalities, thereby gathering critical information necessary for accurate reasoning and response generation. We evaluate our framework on a novel multimodal benchmark dataset, assembled by integrating WebGPT, ScienceQA, AVSD, and ScanQA datasets. Experimental results demonstrate that our CoQ method significantly enhances a foundation model's ability to effectively identify and integrate pertinent sensory information. This leads to improved accuracy, interpretability, and alignment of the reasoning process with diverse multimodal tasks.
\end{abstract}


\section{Introduction}

Recent advancements in large language models (LLMs) have significantly enhanced their reasoning capabilities, primarily through techniques such as Chain-of-Thought (CoT)\cite{Wei22} , which encourage models to explicitly generate intermediate reasoning steps before providing an answer. These methods have markedly improved the interpretability and accuracy of model outputs, particularly for textual reasoning tasks. However, despite these advances, current models predominantly remain limited to unimodal, text-based interactions and often neglect the rich multimodal contexts present in real-world environments.

Human reasoning inherently integrates multiple sensory modalities such as visual, auditory, spatial, and textual to construct coherent interpretations of complex scenarios. For example, when navigating a bustling street, humans simultaneously interpret visual cues from traffic signs, auditory information from vehicle noises, spatial awareness from surrounding structures, and textual instructions from navigation apps. Such comprehensive multimodal reasoning allows humans not only to respond accurately but also to proactively seek missing information by directing attention to relevant sensory channels.

In contrast, existing multimodal language models (MLLMs) typically treat modalities other than text as supplementary inputs, passively incorporating them into their reasoning processes. This passive modality integration constrains the models' ability to dynamically determine what additional sensory information is necessary for comprehending and addressing context-dependent tasks. Consequently, their applicability and effectiveness are significantly diminished in practical, dynamic, real-world scenarios requiring active sensory exploration.

To overcome these limitations, this paper proposes a novel approach Chain of Questions (CoQ) designed explicitly to guide multimodal language models in proactively generating curiosity-driven questions that dynamically identify and engage relevant sensory modalities. This active questioning mechanism enables models to autonomously determine which modalities (vision, audio, spatial perception, etc.) should be activated to gather necessary information from their environment. The CoQ framework thus represents a substantial advancement beyond passive multimodal integration approaches by promoting active, targeted sensory exploration, aligning model reasoning processes more closely with natural human cognition. Our approach introduces a new paradigm of "multimodal curiosity," enabling language models to systematically and selectively query their surroundings, enhancing both the interpretability and accuracy of their multimodal reasoning capabilities.

\subsection{From Prompt to Sensors}

To create more robust AI systems that better mirror human cognitive processes, it is essential to extend reasoning capabilities to actively include multimodal information. This work proposes a novel framework, the Chain of Questions (CoQ), designed explicitly to enhance multimodal reasoning in language models by guiding them to selectively query their environment through curiosity-driven questions. By dynamically generating these questions, the model identifies which sensory modalities are necessary to gather relevant information for solving a given task.

This process is implemented within the framework through four distinct conceptual stages:

$$
\text{Prompt} \rightarrow \text{Question} \rightarrow \text{Task} \rightarrow \text{Sensor}
$$

\begin{itemize}
\item \textbf{Prompt}: The initial textual input provided by the user.
\item \textbf{Question}: Curiosity-driven inquiries that the model formulates to gather relevant multimodal data. A comprehensive list of possible questions is presented in Table 1.
\item \textbf{Task}: Specific operations triggered by these questions, such as face recognition, speech-to-text (STT), or object detection.
\item \textbf{Sensor}: Hardware or software-based modalities activated by tasks, including cameras, microphones, LiDAR sensors, etc.
\end{itemize}

\begin{figure}
  \centering
  \includegraphics[width=\textwidth]{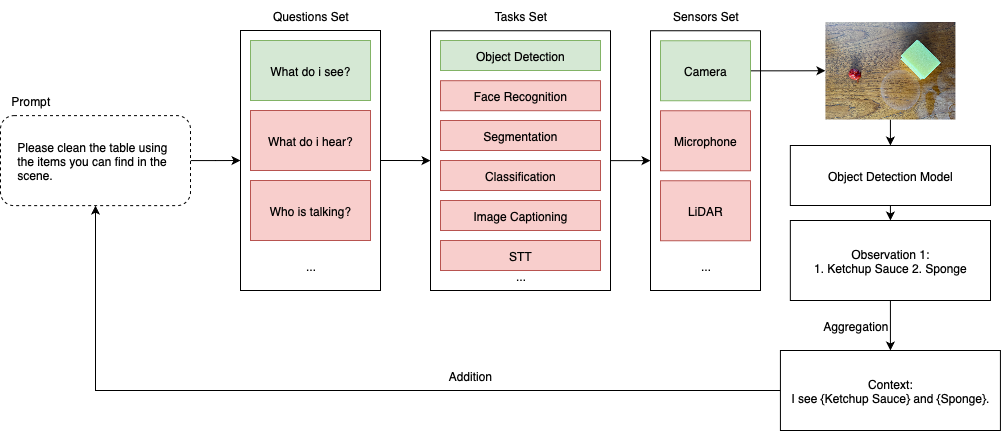}
  \caption{
    Illustration of the Chain of Questions (CoQ) framework for multimodal reasoning. 
    Given a natural language prompt, the model generates a set of curiosity-driven questions, each mapped to specific perceptual tasks (e.g., object detection, speech-to-text). 
    These tasks activate the corresponding sensors (e.g., camera, microphone) to gather environment-specific data. 
    The collected observations are then aggregated into a coherent context, enabling the model to form a structured and grounded response. 
    This process mirrors human-like inquiry and perception, enhancing reasoning through selective multimodal exploration.
  }
  \label{fig:fig1}
\end{figure}

\subsection{Example Workflow}

An illustrative example demonstrating the CoQ framework is shown in Figure 1. Given a user prompt, the model formulates targeted questions about the environment, invokes corresponding tasks, and activates appropriate sensors. The sensor-derived observations are aggregated into a coherent multimodal context, enhancing the model's ability to accurately respond to complex, context-dependent prompts. This stepwise questioning and sensing procedure enables precise and contextually informed reasoning.

\begin{table}
  \centering
  \small
  \begin{tabular}{@{}p{0.6\linewidth}p{0.35\linewidth}@{}}
    \toprule
    Question & Task \\
    \midrule
    What do I see? & Object Detection \\
    Who am I looking at? & Captioning \\
    What are they saying? & STT \\
    What am I hearing? & Sound event detection \\
    What is the sentiment? & Sentiment Analysis \\
    What is the spatial location? & Spatial Detection \\
    What is the pose? & Pose Estimation \\
    What are they doing? & Action Recognition \\
    Who is talking? & Speaker ID \\
    What language? & Language ID \\
    \bottomrule
  \end{tabular}
  
  \caption{Questions and corresponding tasks in the CoQ framework.}
  \label{tab:questions}
\end{table}

\section{Background}

Large Language Models (LLMs) have revolutionized natural language processing (NLP) by enabling systems to understand, generate, and reason with human language at unprecedented scales. Historically, translating complex human problems into formal programming languages was both challenging and resource-intensive ~\cite{Sadiku18, Zheng24}. NLP emerged as a response, developing algorithms for essential tasks such as text classification, summarization, sentiment analysis, and information extraction~\cite{Jin23,Sgroi22}.

Early NLP models, such as Recurrent Neural Networks (RNNs)~\cite{Schmidt19} and Long Short-Term Memory networks (LSTMs)~\cite{Hochreiter97}, were pivotal in sequential language processing but faced scalability and parallelization constraints. The introduction of Transformer architectures with self-attention mechanisms~\cite{Vaswani17} represented a significant advancement, enabling efficient parallel training and the development of large-scale models like GPT~\cite{Radford21}, Gopher~\cite{Rae21}, and Chinchilla~\cite{Hoffmann22}, which demonstrate remarkable linguistic fluency and reasoning capabilities.

Human communication inherently integrates multiple modalities such as text, visuals, audio, and spatial information. Computational multimodality involves effectively representing and processing these diverse data types to enrich understanding and generation tasks. Multimodal Language Models (MLLMs) extend traditional NLP models by integrating these multiple modalities into unified or coordinated representations, significantly improving contextual understanding~\cite{Ngiam11,Arjmand21}.

Prominent MLLMs include CLIP, aligning image and text embeddings for cross-modal retrieval and generation~\cite{Radford21}, ViLBERT, jointly processing visual and textual inputs for tasks such as visual question answering~\cite{Lu19}, and Flamingo, designed for few-shot multimodal interactions~\cite{Alayrac22}. Fusion methods integrate modalities into shared latent representations~\cite{Poria17,Zhang20}, while coordination and fission approaches manage separate yet interrelated modality-specific representations. Additionally, Socratic models leverage predictions from unimodal experts to guide textual reasoning in multimodal contexts without extensive retraining~\cite{Zeng23}.

As language models increase in scale, their emergent reasoning abilities, including few-shot learning and generalization to novel problems, have substantially improved~\cite{Wei22}. However, larger model sizes alone do not guarantee consistent performance on complex reasoning tasks~\cite{Rae21}. Consequently, researchers have introduced various prompting methods designed to enhance model reasoning abilities. Chain-of-Thought prompting explicitly guides models through intermediate reasoning steps, closely mimicking human cognitive processes~\cite{Wei22}. Explanation-based prompting further strengthens model performance by requiring explicit explanations for generated responses, enhancing transparency and clarity~\cite{Lampinen22}. Iterative prompting systematically decomposes complex problems into simpler sub-questions, facilitating structured problem-solving~\cite{Wang22}. Moreover, self-consistency prompting improves reliability by generating multiple candidate solutions and selecting the most internally consistent response, thereby increasing confidence in the final output~\cite{Wang23}.

Building upon the challenges identified in multimodal reasoning, our proposed Chain of Questions (CoQ) framework aims to systematically incorporate diverse modalities into the reasoning process of language models. The central idea behind CoQ is to enable the language model to proactively generate targeted, curiosity-driven questions that explicitly map the required evidence in the user's prompt to corresponding modality-specific tasks. These tasks then activate relevant sensors either hardware-based (e.g., camera, microphone, LiDAR) or software-based (e.g., image recognition, speech-to-text) to collect the necessary observations.

\section{Chain of Questions}

In light of our discussion in the previous section, we talked about the challenges to bring multimodal inference to language models. In our proposed framework we are trying to add the information from other modalities in the reasoning process of the models. The main idea is to motivate the model to ask related questions that map the required evidences in the prompt to specific tasks. Then tasks would use different sensors (Hardwares or Software execution) to collect information about the outer environment of the model. In Table 2 you can see a simple prompt answering that requires the model to see it's outer world.

\subsection{Framework Overview}

The Chain of Questions framework operates through a structured pipeline comprising several interconnected stages. When the model receives a textual prompt (P), it first generates a series of modality-specific questions $(Q = \{ q_1, q_2, \dots, q_k \})$ that clarify what additional multimodal information is required. Each question generated by the model corresponds to a specific task through a task selection function ($\mathcal{T}$), resulting in $T_i = \mathcal{T}(q_i)$. Subsequently, each identified task activates appropriate sensors via a sensor assignment function ($\mathcal{S}$), yielding $S_i = \mathcal{S}(T_i)$. The execution of each task $T_i$ using sensor $S_i$ results in an observation $o_i$.

Once all observations ($\mathcal{O} = \{ o_1, o_2, \dots, o_k \}$) are collected, they are aggregated to form a comprehensive multimodal context (C). The final answer (A) is inferred by integrating this context with the initial user prompt through a reasoning function $F_a$:

$$A = F_a \left( P, \; \text{Aggregate} \left( \{ \text{Execute}(\mathcal{T}(q_i), \mathcal{S}(\mathcal{T}(q_i))) \}_{i=1}^{k} \right) \right)$$

\subsection{Chain of Questions Implementation}

The Chain of Questions framework can be implemented through two primary methodologies: few-shot learning and fine-tuning.

The few-shot learning approach requires minimal resources, as it does not necessitate model retraining. Instead, the model is prompted with carefully designed examples that encourage curiosity-driven questioning to gather relevant multimodal information. This approach is beneficial in resource-constrained environments, allowing quick and efficient deployment.

Alternatively, the fine-tuning approach involves training the foundation model explicitly to generate modality-specific questions during the reasoning process. While this method may incur higher computational costs and resource usage, it potentially offers improved accuracy and consistency in question generation and multimodal reasoning outcomes.

In our experimental evaluation, detailed in the subsequent sections, we primarily adopted the few-shot learning method to illustrate the framework's effectiveness clearly and efficiently.

\section{Dataset}

Due to the novelty of multimodal curiosity-driven reasoning, existing datasets were insufficient for effectively evaluating our proposed Chain of Questions (CoQ) framework. Current datasets predominantly focus on enhancing language model capabilities within single modalities, primarily text-based reasoning. To properly evaluate multimodal curiosity and reasoning, we designed and constructed a comprehensive benchmark dataset by carefully integrating multiple existing datasets representing various modalities.

We combined several specialized datasets, including WebGPT~\cite{Nakano21}, ScienceQA~\cite{Lu22}, AVSD~\cite{Alamri19}, and ScanQA~\cite{Azuma22}, each providing distinct modality contexts. This integration aimed to allow language models to dynamically determine whether additional multimodal information such as visual, auditory, or spatial data is required to answer a given prompt accurately.

The WebGPT~\cite{Nakano21} dataset primarily consists of textual modality, containing 19,578 human-generated prompts initially used to train GPT models. As the dataset exclusively comprises text-based prompts, we marked these instances as not requiring additional multimodal information.

In contrast, the ScienceQA~\cite{Lu22} dataset features prompts with and without supplementary visual evidence. We explicitly divided ScienceQA prompts into two distinct categories: those accompanied by visual evidence (images) and those strictly textual. This classification allowed the language models to recognize prompts explicitly requiring visual input.

Additionally, we employed the AVSD~\cite{Alamri19} dataset to evaluate scenarios involving dialogues within video sequences. In such cases, models must integrate both auditory and visual modalities to comprehend the dialogue context adequately, necessitating the generation of appropriate curiosity-driven questions related to visual observation and auditory understanding.

Finally, the ScanQA~\cite{Azuma22} dataset, which includes 41,363 human-curated question-answer pairs based on 800 ScanNet 3D indoor scans, was incorporated to assess spatial modality comprehension. Prompts from ScanQA explicitly require the model to query spatial information about objects within a given environment.

After meticulous integration and categorization, our final multimodal benchmark dataset consists of 180,629 carefully labeled instances. This comprehensive dataset structure incorporates purely textual prompts, visually supported prompts, audiovisual dialogue prompts, and spatially oriented prompts. This categorization enables rigorous evaluation of a model's capability to effectively identify when additional multimodal information is necessary, thus thoroughly assessing multimodal reasoning performance within the proposed CoQ framework.

\section{Experiment}

This section presents a detailed description of the experimental setup and results, thoroughly evaluating the efficacy of the Chain of Questions (CoQ) framework. The primary goal of these experiments is to assess whether the CoQ method effectively prompts language models to generate appropriate curiosity-driven questions and thereby select suitable multimodal information. Additionally, we explore how this framework performs across models of varying sizes, highlighting its adaptability and robustness.

\subsection{Implementation Details}

Our experiments involved multiple language models to comprehensively evaluate the effectiveness of the CoQ framework across different architectures and sizes. We selected four prominent models: FLAN T5 base (250 million parameters), FLAN T5 large (780 million parameters), FLAN T5 xl (3 billion parameters), and Llama 2 (7 billion parameters). These models represent varying levels of emergent reasoning capabilities, allowing us to examine how the CoQ method scales with model complexity.

Each model was configured with three different decoding strategies: (1) greedy decoding, selecting the token with the highest probability; (2) sampling, promoting diversity in token selection; and (3) beam search, which considers multiple potential outputs simultaneously, although with increased computational overhead. Sampling and beam search configurations particularly depend on the flexibility and robustness of the function mapping generated questions to corresponding multimodal tasks.

\begin{table}[h!]
\centering

    \begin{tabular}{|l|c|c|}
    \hline
    \textbf{Model} & \textbf{Parameters} & \textbf{Type} \\
    \hline
    FLAN T5 base  & 250 million & encoder/decoder \\
    \hline
    FLAN T5 large & 780 million & encoder/decoder \\
    \hline
    FLAN T5 xl    & 3B           & encoder/decoder \\
    \hline
    Llama 2       & 7B           & decoder only \\
    \hline
    \end{tabular}
        \caption{Models used in our experiments with their parameter counts and architectural types.}
    \label{tab:models}
\end{table}

Inference was performed using specialized GPU hardware P100 GPUs for FLAN T5 models and A100 GPUs for Llama 2 to ensure efficient computational performance. Given the dataset's substantial size, batching methods were employed during data loading and inference processes to optimize efficiency and resource usage.

To maintain computational efficiency and clarity in evaluating the CoQ framework, we primarily utilized a few-shot learning approach. Specifically, the models were prompted with explicit instructions and illustrative examples designed to elicit curiosity-driven multimodal questions relevant to each prompt.

The primary focus of these experiments was not on final answer generation, but rather on evaluating the accuracy and relevance of the questions generated by the models.

\subsection{Experimental Results and Analysis}

Our experimental results indicate substantial differences in model performance in generating relevant curiosity-driven multimodal questions, strongly correlated with model size and architectural design. The key results of our experiments are summarized in Tables 3 and 4.

\begin{table}[h!]
\centering
\small
\begin{tabular}{|l|r|r|c|}
\hline
\textbf{Model} & \textbf{Match} & \textbf{Mismatch} & \textbf{Match \%} \\
\hline
LLaMA 7B & 79,355 & 101,274 & 43.9\% \\
\hline
FLAN T5 XL & 137,701 & 42,928 & 76.2\% \\
\hline
FLAN T5 Large & 47,511 & 133,118 & 26.3\% \\
\hline
FLAN T5 Base & 31,861 & 148,768 & 17.6\% \\
\hline
\end{tabular}
\caption{
Comparison of matched versus mismatched outputs across different model variants. 
This table reflects how accurately each model aligned its response with the correct modality after generating curiosity-driven questions. 
Higher match percentages indicate better reasoning alignment and task understanding.
}
\label{tab:comparison-results}
\end{table}

Table 3 clearly demonstrates that the FLAN T5 xl model (3 billion parameters) achieved the highest accuracy in producing relevant multimodal questions directly aligned with the input prompts. In contrast, smaller models such as FLAN T5 base and FLAN T5 large showed limited effectiveness, often failing to generate appropriately targeted questions. Llama 2 (7 billion parameters), despite its larger size, underperformed relative to FLAN T5 xl, potentially due to its decoder-only architecture, which favors more diverse but less precise token generation.

\begin{table}[h!]
\centering
\begin{tabular}{|l|r|r|c|}
\hline
\textbf{Model} & \textbf{Asked} & \textbf{Did Not Ask} & \textbf{Asked (\%)} \\
\hline
LLaMA 7B & 74,036 & 106,593 & 41.0\% \\
\hline
FLAN T5 XL & 144,547 & 36,082 & 80.0\% \\
\hline
FLAN T5 Large & 130,941 & 49,688 & 72.5\% \\
\hline
FLAN T5 Base & 60,773 & 119,856 & 33.6\% \\
\hline
\end{tabular}
\caption{
Rate of modality-related questions generated by each model as a proxy for curiosity. 
This table captures how often each model attempted to engage in curiosity-driven exploration by asking questions, regardless of their correctness. 
Higher asking rates suggest greater inherent curiosity or responsiveness to the CoQ prompting framework.
}
\label{tab:question-generation-results}
\end{table}

Furthermore, Table 4 illustrates the general curiosity exhibited by each model, capturing instances where any type of modality-related question was produced, irrespective of accuracy. Here, FLAN T5 models consistently displayed higher overall curiosity about their environment compared to Llama 2. Notably, FLAN T5 large frequently generated multimodal queries, albeit often less targeted or relevant to the specific prompts.

These findings emphasize the significant influence of both model architecture and parameter scale on the successful implementation of the CoQ framework. Additionally, they highlight the importance of refining prompting strategies and improving task-mapping functions to enhance both the relevance and precision of curiosity-driven questioning.

Overall, our experimental analysis confirms the effectiveness of the Chain of Questions framework in guiding multimodal curiosity and reasoning systematically. These results suggest promising avenues for further enhancements through advanced fine-tuning methods, optimized prompts, and deployment of larger, more sophisticated model architectures. Ultimately, the CoQ framework represents a significant step towards creating more sophisticated, contextually aware language models capable of effectively operating in real-world environments.

\bibliographystyle{unsrt}  
\bibliography{main}  

\end{document}